\documentclass[preprint,12pt]{elsarticle}

\usepackage{amssymb}
\usepackage{url}
\usepackage{booktabs} 
\usepackage{hyperref}
\usepackage{caption}
\usepackage{rotating}
\usepackage{graphicx}
\usepackage{multirow}
\usepackage{array}
\usepackage{amsmath}
\begin{document}

\begin{frontmatter}


% \author{Name\corref{cor1}\fnref{label2}}

% \author{Anonymous Author(s)}

\author[1]{Muhammad Raees}
% \author[2]{Inge Meijerink\fnref{label2}}
% \cortext[cor1]{Muhammad Raees (mr2714@rit.edu)}
% \cormark[1]
\ead{raees.se@must.edu.pk}
\author[1]{Samina Fazilat}
\ead{samina.se@must.edu.pk}
\affiliation[1]{organization={Mirpur University of Science and Technology},
            city={Mirpur},
%          citysep={}, % Uncomment if no comma needed between city and postcode
            state={AJK},
            country={Pakistan}}

\title{Lexicon-Based Sentiment Analysis on Text Polarities with Evaluation of Classification Models}

\begin{abstract}
Sentiment analysis possesses the potential of diverse applicability on digital platforms. Sentiment analysis extracts the polarity to understand the intensity and subjectivity in the text. This work uses a lexicon-based method to perform sentiment analysis and shows an evaluation of classification models trained over textual data. The lexicon-based methods identify the intensity of emotion and subjectivity at word levels. The categorization identifies the informative words inside a text and specifies the quantitative ranking of the polarity of words. This work is based on a multi-class problem of text being labeled as positive, negative, or neutral. Twitter sentiment dataset containing 1.6 million unprocessed tweets is used with lexicon-based methods like Text Blob and Vader Sentiment to introduce the neutrality measure on text. The analysis of lexicons shows how the word count and the intensity classify the text. A comparative analysis of machine learning models, Naïve Bayes, Support Vector Machines, Multinomial Logistic Regression, Random Forest, and Extreme Gradient (XG) Boost performed across multiple performance metrics. The best estimations are achieved through Random Forest with an accuracy score of 81\%. Additionally, sentiment analysis is applied for a personality judgment case against a Twitter profile based on online activity. 
\end{abstract}


\end{frontmatter}

% \linenumbers


\section{Introduction}
Social networks have seen ever-increasing growth with the shift of news media to online networks producing a copious quantity of information \cite{bialik28}. 
Leveraging social media capabilities to capture people’s ideas, expressions, and emotions on various topics is an interesting task. 
Social networks provide capabilities to form narratives within the network and between them. Such capabilities have revolutionized the information flow and the ways it influences the network by shaping people’s opinions \cite{gorodnichenko2021social29}. 
The influence of opinion is instinctive as often people make decisions based on others’ opinions (e.g., family, friends). However, opinions and narratives engendered through social media can have a huge impact. The study to generate insights out of expressed opinions is called sentiment analysis. The analysis can extract sentiments and emotions from user data that can lead to meaningful conclusions. Analysis can investigate human feelings, thoughts, behaviors, or experiences. The terms are often unified as opinion mining \cite{liu2012sentiment31}. 

Analysis of opinions and narratives is tedious due to the complexity of language and the diversity of expressions. The problem is exacerbated by having a variety of contextual scenarios essentially making the task more analytical. Opinions are also complex and often of two types: direct and comparative opinions \cite{liu2010sentiment2}. Direct opinions are often explicit and explain a specific entity while comparative opinions involve more than one entity and a degree of comparison. Such differentiation is necessary, for example, a direct opinion is useful to understand the client’s confidence in a product or service. Identifying a potential political candidate may involve a comparative opinion. Nowadays, online networks (e.g., Twitter) are the main platforms where people opine about specific topics \cite{sun2014analyzing30}. The components of text in a tweet (a Twitter post) provide useful aspects about the author, the opinion, and the way the opinion is expressed \cite{osimo2012research1}. Most of the time, the text conveys a connotation or an attitude that can be categorized into three different dimensions or orientations: Positive, Negative, or Neutral. Orientations have been based on the estimates of the polarity. The polarity shows the strength of the opinion or attitude of the text. It is essential to understand, observe, and verify the attitude of the author towards the object or subject in question. This methodology helps to identify the subjectivity or objectivity of a sentence. An objective statement or words do not necessarily provide the degree of an attitude. Therefore, removing objective lexicons before polarity often helps in improving the performance of the process \cite{aue2005customizing4, khanintelligent}.

The understanding of the language is critical in sentiments. For instance, a sardonic expression might give a false context impression. The reliability of the content and recognition of the opinion holder is crucial to avoid fake or unreliable data. Lexicon-based methods eliminate the probability of uninformative expressions and data unreliability by focusing on the polarity and subjectivity of words. A vast amount of work is concentrated on sentiment analysis dealing with positive and negative polarities. However, neutral polarity has an important role, for instance, neutral polarity can help to eliminate irrelevant product reviews from consideration or to shape an opinion of personality. This work concentrates on the application of lexicon-based sentiments with an evaluation of machine learning algorithms with the following contributions. 
\begin{enumerate}
  \item Performing exploratory data analysis to elicit insights about the polarity and subjectivity.
  \item Transformation, inclusion, and study on the impact of lexicon neutrality in sentiments.
  \item Evaluating the performance of machine learning algorithms to provide a guideline for model selection for sentiment analysis task (with an example study to perform personality judgment based on opinions expressed on Twitter).
\end{enumerate}

This study will demonstrate the methodology that uses all polarity values (positive, negative, neutral) to perform sentiment analysis and compare the performance of machine learning models trained on sentiment datasets. This section provided an overview of the sentiment analysis and associated issues with it. The rest of the paper is organized as follows: Section 2 highlights some related studies and background information. Section 3 explains the process of experiments and methods to generate polarities. The results of the experiments and discussions are presented in Section 4. The final section concludes the paper.

\section{Background and Related Work}
\label{sec:relatedwork}
Machine learning techniques are being applied in all sorts of domains and practical contexts \cite{raees2024explainable}.
Approaches for textual analysis are often categorized into keyword recognition, lexical affinity, and conception-level methods \cite{cambria2013new3}. Machine learning algorithms, producing superior results, have been highly useful in sentiment analysis applications \cite{hasan2018machine9,kundi2014lexicon10, asghar2017lexicon11}. However, there is little support for benchmark model comparison and evaluation. 
Ensemble methods have performed relatively better than individual models \cite{hu2004mining18}. However, some studies have reported individual models like Naïve Bayes, Artificial Neural Networks, K-Nearest Neighbors, and Support Vector Machines being applied for sentiment classification \cite{califf2003bottom21, choi2005identifying22}. Other studies use lexicon-based methods using parts of speech (POS) to extract sentiment \cite{read2005using15, agarwal2011sentiment16}. 
Sentiments can also be evaluated in multiple phases by identifying important contextual features in each phase and thus annotating them with the polarity. Simple exploratory analysis can show the sentiments through the examination of opinions and identifying the ratio of positive and negative words \cite{pang2002thumbs26,khanintelligent36, khanintelligent}.
A root-cause analysis can also provide an effective strategy to understand what to build in models \cite{raees2020study}.

Words, phrases, or sentences in the documents serve as keywords to identify the semantic orientation of the text. Sentiment analysis can be performed at four disparate granularity: document level, sentence level, word level, and feature level. Document-level analysis presumes the content has a single opinion holder. Pang and Lee \citep{pang2005seeing32} developed a score rating method using various classifiers. The document-level analysis is often complemented by sentence analysis due to the nature of the subjectivity involved. A lexicon or corpus-based sentiment analysis is often performed on the word level. The subjectivity and association of words are calculated with assistive lexical relation \cite{kamps2004using33}. Corpus-based mostly relies on syntax and mathematical methods with word associations \cite{turney2002thumbs34}. The feature-based approach focuses more on extracting features by identifying parts of sentences such as verbs, nouns, and phrases \cite{popescu2007extracting35}. Hence, feature and word-level approaches are more suitable for effective analysis. Twitter, like various online platforms, data is an excellent corpus for analysis. Abundantly, the uninformative and irrelevant information in the corpus requires preprocessing before analysis. The effects of preprocessing on data collected from Twitter are highlighted in various studies \cite{krouska2016effect5}. Eliminating unnecessary lexicons and features can also be classified as preprocessing.

Various types of problems are embarked upon with sentiment analysis. A prototype for customer reviews \cite{sarlan2014twitter7} classifies tweets into positive and negative classes showing a similar application. An ensemble approach in comparison with other classification methods is applied to airline service data. Such sentiment analysis applications gather reviews of customers expressed in negative or positive polarities based on their experience with the product or service. Studies \cite{gorodnichenko2021social29} examine the implications of opinion mining during the U.S. elections in 2012 and 2016 for presidential candidates. Authors consider various sources including real-time evaluation of sentiment on Twitter. Studies like these are useful for political figures to target future perspectives of public opinion and the electoral process.

\section{Polarity Analysis Methodology}
\label{sec:methodology}

Various domains analyze sentiments with specialized applications to identify useful information. Historically, user reviews have served as a primary service feedback source. The automated classification of reviews into their respective polarity requires capturing polarity and subjectivity. Fake or irrelevant reviews may jeopardize the classification process without the annotation of polarity in opinions. This section outlines an end-to-end sentiment analysis process applied to a Twitter dataset. The preprocessing eliminates the meaningless words from the text and provides the most useful features for further analysis. Polarity generation is achieved through two disparate methods. Useful features are extracted after the polarity generation and various models are trained. The machine learning models with lower complexity can be trained to achieve good results for future analysis. Figure 1 depicts an overview of the methodology being followed for the whole process.

\begin{figure*}
  \centering
  % note that \textwidth is used instead of \linewidth
  % This ensures that the graphics width is 60% of the "page" (text block), and not just 60% of the current text column
  % See https://tex.stackexchange.com/a/17085/9075 for details
  \includegraphics[width=.8\textwidth]{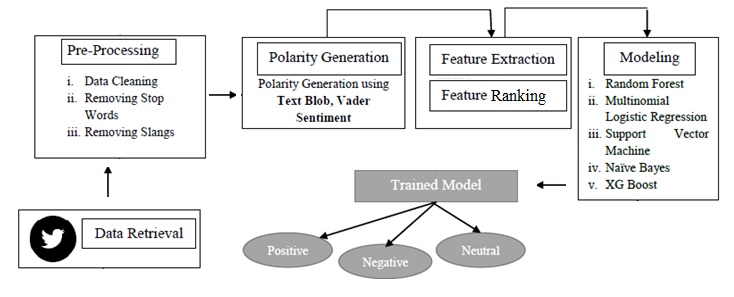}
  \caption{Overview of the basic methodology}
  \label{fig:methods}
\end{figure*}

\subsection{Data Retrieval and Pre-processing} 
Data pre-processing is an integral component in natural language processing because it transforms the data so that it can be further used for analysis. In textual corpus, preprocessing relevant features and proper representation affects model accuracy \cite{soldatova2014exact238}. The data is retrieved from the Twitter sentiment 140 \cite{go2009twitter37} platform. The platform allows the discovery of different sentiment analyses of brands, products, or topics.  Initially, the raw dataset contained 1.6 million unprocessed tweets with tantamount negative and positive polarities, respectively. Pre-processing steps include non-conforming data cleaning through regular expressions for tweets. Numbers, URLs, Twitter usernames, special characters, along other non-essential tokens are removed in this step. Regular expressions are generated to remove patterns in tweets having such type of data. Secondly, some words do not necessarily provide any useful meaning. These are usual words called stop words in English (like any other language) and focus only on relevant information instead of the commonly used language. Removal of such words is important so that the actual content can be derived from the sentence. These mostly conform to connecting words and phony phrases. The natural language token library is comprised of a corpus module that provides objects and functions helpful in removing the stop words. Separately, a list of slang is defined for possible abbreviations that are expected to be present in the dataset. Some cases require lemmatizing or stemming the original words to exact words where the meaning can be captured. The lemmatization produces a derivative of the original word to ensure the correct meaning is derived from a dictionary, instead of its derivation. Stemming, on the other hand, reduces the word to its basis by truncating the unnecessary word endings. The final stage of preprocessing eliminates the slang from the dataset by finding and replacing them with their complete and understandable form of words.

\subsection{Polarity Generation}
Polarity generation is a very important step to identify the type of tweets. For a better understanding of the dataset and reclassification of negative and positive tweets after the application of preprocessing, the polarity of tweets is recalculated with Text Blob and Vader Sentiment. Both methods are based on lexicon-based techniques and produce a mapping between words and sentiments. Text Blob is very useful for sentiment analysis and provides two metrics of textual data: polarity and subjectivity. Polarity output ranges from -1 to 1, where negative shows the negative polarity and vice versa. To eliminate the reciprocity of negativity or positivity unnecessarily, a score closer to 0 will be classified as neutral. Subjectivity score ranges from 0 to 1 referring to opinion or judgment. The Vader Sentiment technique was also used to find a polarity score, which further helped determine the tweet's class. In this case, if the Vader score was greater than 0.05 then the tweet was positive and if the score was less than -0.05 then the score was negative. Again, the score between this range will be used for neutral tweets. Figure 2 shows the evaluation of both the sentiment scores on the dataset after pre-processing. The pre-processing eliminates some unnecessary information to have variations in the output labels. The dataset originally contained an equal number of positive and negative tweets. The figure also shows the impact of neutrality employed with Text Blob and Vader Sentiment as some tweets are classified as neutral from the set of positive or negative ones. With the introduction of neutral tweets, the distribution of tweets in three categories can also be visualized. The Text Blob shows a highly aberrant distribution with a significant reduction of the tweets originally marked as negative. Both methods employ the lexicon-based method to extract polarity and subjectivity. However, Vader is more sophisticated for social media analysis and thus identifies subtle idiosyncrasies of the content that appears on social media in comparison to Text Blob. Also, the Vader has a smaller range of scores (i.e., -0.05 to 0.05) that could differentiate the results more than the Text Blob. Therefore, the Vader Sentiment is a better polarity measure to be considered for further processing.

\begin{figure*}
  \centering
  % note that \textwidth is used instead of \linewidth
  % This ensures that the graphics width is 60% of the "page" (text block), and not just 60% of the current text column
  % See https://tex.stackexchange.com/a/17085/9075 for details
  \includegraphics[width=.9\textwidth]{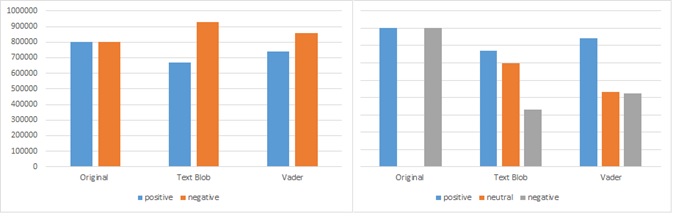}
  \caption{The comparison of transitioning of the original dataset (left – containing only the positive and negative tweets) along with the modified dataset (right – to include neutral tweets) with the Text Blob and Vader Sentiment.}
  \label{fig:tweets}
\end{figure*}

\subsection{Feature Selection and Ranking}
For feature extraction, a simple method, the term frequency-inverse document frequency (TFIDF) is employed. The method magnifies the importance of words in the provided text corpus and each feature represents the frequency of words appearing. It helps to identify relevant word features that can be utilized for the approximation of text polarity using machine learning models. These features are associated with feelings being conveyed in a set of words. Visualizations of common words shown in the next section report the reason for the sentiment polarity, for example negative, positive, or neutral. The ranking has been performed to identify the meaningful words in the text to remove the words with lower polarity and subjectivity. The removal of unnecessary features is very similar to the previous pre-processing step. However, feature ranking only removes the words that have some level of impact but not too much. The difference is controlled by employing the Jaccard similarity measure. The threshold varies according to the size of the input text with a random selection of threshold between 60\%-80\% of similarity ensured between the original and truncated text so that most of the useful information is not lost. Hence, this greatly reduces the input set and ranks the features according to their impact on the polarity and subjectivity. This step also reduces the computations needed for the machine learning models making them simpler and faster to run.

\subsection{Training Models}
Lexicon methods can be completed with machine learning to build and train models. The following machine learning models are chosen due to their simplicity of explanation and lower computational power. These models are trained and tested on arbitrary portions of datasets to find optimum parameters and models. For this measure, various data splits are tried for evaluation and reporting. To build an approximation of the input data various classification models are used for the classification of positive and negative tweets. A short description of the models applied is explained as follows.

a) Naïve Bayes: It is a simplistic model based on Bayes’s assumption of conditional probabilities. It performs calculations on the fly and in real-time for predicting the output of an example. Naïve Bayes uses a probabilistic theorem to estimate the probabilities of the target class. The output of data is the highest probability calculated across a set of output classes. Naïve Bayes is based on simple assumptions, however, it performs exceptionally well even in complex scenarios.

b)	Multinomial Logistic Regression: The second class of algorithms used is Multinomial Logistic Regression (MLR) which is quite similar to Binary Logistic Regression where the class labels are only positive and negative. Multinomial Logistic Regression is used for classification in cases where there are more than two output classes. Similar to logistic regression, it calculates the maximum likely hood finding the probability of any relationship to a specific class.

c)	SVM: Support Vector Machines (SVM) are versatile sets of algorithms used in supervised learning for classification and regression problems.  SVMs work by using a hyperplane to separate the classes in case of classification. If we are provided with the labeled data then considering a two-dimensional plane, the hyperplane will be a line separating the plane into two parts so that each class is separated on the side of the hyperplane.

d)	Random Forest: It is a classic example of Classification and Regression Tree (CART) models. The model is an ensemble of various separately trained models to enhance predictive power. In comparison to building a fairly complex model, it is convenient to build a set of simple models with varying parameters. The random forest classifier makes multiple decision trees on different variables and then combines the result. Each tree is independent in decision making and predicts the target variable for each example. The final result is obtained from the majority voting of each decision tree.

e)	XG-Boost: Extreme Gradient (XG) Boost is an algorithm that works based on boosting where it attempts to convert the weak learners to strong learners by assigning weights to them. All the weak learners combine to form strong learners. This is also a classic example of ensemble methods. It is slightly different from the random forest in terms of the construction of decision trees. Random forest integrates the fully functional decision trees, while the XG-Boost develops smaller trees and aggregates/boosts their predictions. 

\section{Results and Discussion}
\label{sec:results}

Models have been evaluated on metrics of accuracy, precision, recall, and F-1 scores for reporting of results for model evaluation. Accuracy defines the measure of correct predictions over total predictions made by models. Precision refers to the correctness of the models and how confidently a positive prediction stands over total positive predictions. Recall checks the true positive rate over the positive classes themselves. In various cases, precision and recall individually do not provide enough ground to report the performance of the model. Therefore, the combination of precision and recall (called the F-1 Score) is the preferred performance measure. The reported performance results vary according to the splitting of data into training and testing sets. The dataset includes 1600000 processed no-emoticon tweets for training and testing of the models. The Text Blob and Vader Score analysis slightly modifies the original dataset as described in the previous section.

\subsection{Exploratory Analysis}
Exploratory methods are used across domains to support data interpretations to group that together with expert understanding \cite{raees2023four}.
The exploratory data analysis provides complementary evidence about the predictions of the models and supports the findings of the process. 
The analysis corroborates that data removal through necessary preprocessing and feature extraction still provides meaningful information observed through the most common words between both data attributes. Table 1 shows an analysis of common words found in the text to identify as positive, negative, and neutral respectively. The analysis just shows the commonality of words in tweets that are classified. Many other words having more stronger polarity might not be abundant in the dataset but could affect the label of a tweet more strongly. Similarly, Table 2 shows the occurrences of unique words that contribute to the polarity of tweets. This shows a better correlation between the polarity and the words appearing in the tweet. However, this provides a bird-eye view of how lexicon-based methods use the polarity of words to classify the corresponding tweets into their respective category.

\begin{table}
 \caption{The lexicon-based occurrence of common words in positive, negative, and neutral tweets} 
 \label{tab:maxcount}
  \centering
  \begin{tabular}{l|l|l|l|l|l|l}
    \toprule
    Sr.	& Positive & Count & Negative & Count & Neutral & Count \\
    \midrule
    1	& thanks & 3801896 & sorry & 96884 & goodnight & 167626 \\
2 & Thank & 1696695 & I’m & 88759 & work & 120432 \\
3 & Good & 1395890 & hurts & 86270 & headache & 105030 \\
4 & Day & 1258270 & wrong & 54765 & hey & 63444 \\
5 & Get & 1164261 & tummy & 52155 & morning & 46227 \\
6 & everyone & 1148290 & sucks & 51914 & I’m & 37678 \\
7 & followers & 1120029 & what’s & 51626 & back & 37533 \\
8 & Add & 1118448 & died  & 47367 & cant & 36252 \\
    \bottomrule
  \end{tabular}
\end{table}

\begin{table}
  \caption{The lexicon-based occurrence of unique words in positive, negative, and neutral tweets}
  \label{tab:uniquecount}
  \centering
  \begin{tabular}{l|l|l|l|l|l|l}
    \toprule
    Sr.	& Positive & Count & Negative & Count & Neutral & Count \\
    \midrule
1 & Best & 99722	& dangerous & 44660 & think & 41627 \\
2 & Joy & 86038	& fake	& 37576 & support & 39551 \\
3 & welcome & 84249 & cutting & 35036 & sports & 39393 \\
4 & special & 71363 & violent & 33674 & update & 38180 \\
5 & excellent & 64597	& grief & 27406 & twitter & 32981 \\
6 & amazing & 52369 & sick & 21670 & followers & 31375 \\
7 & peaceful & 33752 & desperate & 21163 & morning & 30373 \\
8 & Worth & 29890 & outrageous & 18239 & find & 28114 \\
    \bottomrule
  \end{tabular}
\end{table}

\begin{table}
  \caption{Precision (p), Recall (R), and F-1 (F) scores on 70-30 data split}
  \label{tab:freq}
  \centering
  \begin{tabular}{lccc}
    \toprule
    Model&60-40&70-30&80-20\\
    \midrule
    Random Forest & 79\% & 81\% & 79\% \\
    Multinomial Logistic Regression & 76\% & 78\% & 77\% \\
    Support Vector Machines & 69\% & 79\% & 78\% \\
    Naïve Bayes & 61\% & 58\% & 57\% \\
    XG Boost & 73\% & 78\% & 78\% \\
  \bottomrule
\end{tabular}
\end{table}

\begin{table}
  \caption{Precision (P), Recall (R), and F-1 (F) scores 70-30 data split for positive, neutral, and negative tweets (left-right)}
  \label{tab:freq}
  \scriptsize
  \begin{tabular}{llllllllll}
    \toprule
    Model&Pos-P&Pos-R&Pos-F&Neu-P&Neu-R&Neu-F&Neu-P&Neu-R&Neu-F\\
    \midrule
    Random Forest&85\%&89\%&87\%&65\%&79\% & 71\% & 84\% & 63\% & 72\% \\
    ML-Regression&78\%&93\%&85\%&83\%&7\% & 13\% & 80\% & 69\% & 74\% \\
    SVM&84\%&88\%&86\%&57\%&42\%&49\% & 75\% & 72\% & 73\% \\
    Naïve Bayes&72\%&3\%&67\%&46\%&29\% & 35\% & 43\% & 58\% & 50\% \\
    XG Boost&83\%&84\%&84\%&61\%&66\% & 63\% & 71\% & 67\% & 69\% \\
  \bottomrule
\end{tabular}
\end{table}

\subsection{Model Training Results}
Comparing models with different settings is an effective strategy to study the impact of which works better for a particular case \cite{manzoor2022ancient}.
For this study, the models are run on various splits for better estimation. The estimations of the models also corroborated with 5-fold cross-validation for each split to provide a generalization of performance measures. The final split for the assessment of the performance of models is the 70-30 ratio with cross-validation. The results of all five models run on the 60-40, 70-30, and 80-20 ratios are presented in Table 3. Random Forests outperform other models with an accuracy of 81\%. At the same sampling, Naïve Bayes provides the lowest accuracy of 58\%. The remaining models have almost similar performance scores. Therefore, the remaining experiments are performed on the 70-30 data splitting. Conclusively, the performance of random forest in the classification of tweets is better than the other models in all the data splits. The randomness in generating trees in a forest provides a substantial indicator for slightly better results. Table 4 shows the results of other performance criteria e.g., Precision (P), Recall (R), and F-1 Score (F) on the participating models using the same dataset.

Here, the table outlines the results of classification models for positive, negative, and neutral tweets. Analytically evaluating the results shows that the utmost precision of 85\% was achieved by Random Forest whereas the least is 72\% in Naïve Bayes for positive tweets. On the same data, a recall of 93\% is achieved by Multinomial Logistic Regression whereas the lowest is 63\% in Naïve Bayes for positive tweets. Likewise, Random Forest shows an F1-Score of 87\% whereas the least is 67\% in Naïve Bayes for positive tweets. For Neutral Tweets, a precision of 83\% in Multinomial Logistic Regression is observed whereas the least is 46\% in Naïve Bayes. A recall of 79\% in Random Forest against the least surprisingly 7\% in Multinomial Logistic Regression for neutral tweets. F1-Score of 71\% in Random Forest is achieved whereas the least 13\% in Multinomial Logistic Regression for neutral tweets. For Negative Tweets, the best precision of 84\% is through Random Forest whereas the least is 43\% in Naïve Bayes. Support Vector Machine estimate recall of 72\% whereas the least is 58\% in Naïve Bayes for negative tweets. F1-Score reports 72\% by Random Forest whereas the least 50\% in Naïve Bayes.  Conclusively, the estimations produced by random forests are slightly better than the rest of the models. The Naive Bayes, being the simplest of all, does not accurately capture the underlying patterns most of the time and thus performs poorly of all the methods. Overall, the performance still needs further improvements in the other included models. This word does not aim to include more complex models due to the nature of the problem and the computational costs associated with that. However, future work could include some optimized complex models like neural networks and deep neural networks. The possible future direction to enhance performance score can also be looking at bettering preprocessing methods as well as enhancing the efficacy of feature selection.

\subsection{Personality Assessment}
For the second set of experiments, the analysis is carried out for opinion mining in the application of personality judgment. The end-to-end sentiment analysis process using Text Blob and Vader Sentiment is employed to calculate the polarity of tweets. For the proposed methodology tweets of a renowned politician are accessed and input as a dataset. Around 3000 tweets were fetched from Tweepy. Tweepy is an open-source library available on GitHub which is used to access Twitter data via OAuth authentication. Figure 3 shows the sentiment classification of tweets using both methods on the dataset after the essential preprocessing. A similar data analysis again shows the polarity of the tweets and how the connotation of words causes the lexicon analyzer to calculate the polarity. The results of the most frequent words and the unique words in positive, negative, and neutral tweets are illustrated in Table 4. The analysis depicts common words in tweets to show the polarity. The unique word count shows the count of unique words occurring in different tweets marked as positive, negative, and neutral.

Due to the limited size of the dataset, the frequency of the common and unique words decreases proportionally. The results indicate that many of the tweets are positive. The negative and neutral tweets ratio is comparatively low from Positive tweets. The ratio of negative tweets with positive tweets combined with neutral tweets further decreases to form better estimates of personality. Results are shown according to the accuracies and confusion matrix. Here, the trained model random forest is used for reporting results while testing the dataset. Table 5 outlines the confusion matrix showing the TPs, TNs, FPs, and FNs for all three classes i.e., Positive, Neutral, and Negative classes.

\begin{table}
  \caption{The lexicon-based occurrence of common and unique words (Cnt) in positive (P), negative (Ng), and neutral (Nt) tweets}
  \label{tab:freq}
  \scriptsize
  \setlength\tabcolsep{4pt}
  \begin{tabular}{llllllllllllll}
    \toprule
    Sr.&PWord&Cnt&PUnique&Cnt&NgWord&Cnt&NgUnique&Cnt&NtWord&Cnt&NtUnique&Cnt\\
    \midrule
    1 & great & 777 & birthday & 22 & Fake & 173 & killing & 11 & Trump & 23 & sept & 4 \\
    2 & amp & 268 & amazing & 18 & News & 169 & fraud & 10 & Amp & 23 & prolife & 4 \\
    3 & USA & 255 & honored & 17 & People & 140 & dumb & 8 & Jobs & 20 & weekly & 4 \\
    4 & thank & 250 & champ & 15 & Democrats & 111 & dirty & 8 & USA & 18 & nasdaq & 3 \\
    5 & people & 219 & freedom & 14 & Trump & 96 & dumbest & 6 & Time & 17 & belgium & 2 \\
    6 & us & 186 & behalf & 14 & Us & 95 & ross & 6 & 14 & yom & 2 \\
    7 & make & 182 & kevin & 12 & Country & 94 & spying & 6 & Day & 14 & greek & 2 \\
    8 & big & 167 & sign & 12 & Many & 93 & leak & 5 & Big & 14 & proto & 2 \\
  \bottomrule
\end{tabular}
\end{table}

\begin{table}
  \caption{Accuracy results in personality judgment data using random forest}
  \label{tab:freq}
  \centering
  \begin{tabular}{lccc}
    \toprule
    Actual/Predicted & Positive & Neutral & Negative\\
    \midrule
    Positive & 133 & 22 & 56 \\
    Neutral & 1 & 99 & 26 \\
    Negative & 25 & 32 & 466 \\
  \bottomrule
\end{tabular}
\end{table}

\begin{figure}
  \centering
  \includegraphics[height=5cm]{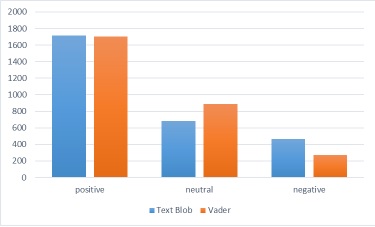}
  \caption{Tweets Polarities division with TextBlob and Vader Sentiment}
  \label{fig:judge}
\end{figure}

\section{Conclusion}
\label{sec:conclusion}
The work showcases a multi-classification problem with an evaluation of the performance of various machine learning models on a lexicon-based text analysis method. Lexicon-based methods are utilized for the analysis of sentiments on datasets to gain insights into the opinions of the people. The analysis insights can be utilized for various purposes e.g., product reviews, political agenda, or a topic discussion. Here, an exploratory analysis has been carried out to highlight the significance of the lexicons on polarities. Exploratory analysis and experimental results indicate that only selected features (words) have a better impact on predictions. The method shows the use of a lexicon analysis through Text Blob and Vader Sentiment techniques before applying feature engineering to optimize learning processes. The study shows a comparative analysis of five simple and fast machine-learning models using various splits. The random forest model achieves the best results with an accuracy of 81\%. The study also forms a rudimentary personality assessment on the tweets derived from Twitter using an open-source library Tweepy (forming a different dataset) to extract opinions and make assessments. The work can be extended by looking at the methods to better feature selection, ranking, and preprocessing.

 \bibliographystyle{elsarticle-num} 
 \bibliography{cas-refs}





% \end{thebibliography}
\end{document}